\documentclass{article} 
\usepackage{iclr2018_conference,times}
\usepackage{hyperref}
\usepackage{url}
\usepackage{bbm}
\usepackage{amsmath,amssymb}
\usepackage{bm}
\usepackage{caption}
\usepackage{dsfont}

\usepackage{hhline}
\usepackage{subcaption}
\usepackage{mathtools}
\usepackage{wrapfig}
\usepackage{color,soul}
\usepackage{multirow}

\usepackage{algorithm}
\usepackage{algpseudocode}

\usepackage{xcolor}
\hypersetup{
    colorlinks,
    linkcolor={red!50!black},
    citecolor={blue!50!black},
    urlcolor={blue!80!black}
}

\title{
Hierarchical and Interpretable Skill Acquisition in Multi-task Reinforcement Learning}

\author{Tianmin Shu\thanks{This work was done when the author was an intern at Salesforce Research.} \\
University of California, Los Angeles\\
\texttt{tianmin.shu@ucla.edu} \\
\And
Caiming Xiong\thanks{Corresponding author}  \& Richard Socher \\
Salesforce Research \\
\texttt{\{cxiong, rsocher\}@salesforce.com} \\
}

\algnewcommand\INPUT{\item[\algorithmicinput]}
\algnewcommand\algorithmicinput{\textbf{Input:}}
\algnewcommand\OUTPUT{\item[\algorithmicoutput]}
\algnewcommand\algorithmicoutput{\textbf{Output:}}

\iclrfinalcopy 

\begin{document}

\maketitle

\begin{abstract}
Learning policies for complex tasks that require multiple different skills is a major challenge in reinforcement learning (RL). It is also a requirement for its deployment in real-world scenarios.
This paper proposes a novel framework for efficient multi-task reinforcement learning. 
Our framework trains agents to employ hierarchical policies that decide when to use a previously learned policy and when to learn a new skill. This enables agents to continually acquire new skills during different stages of training.
Each learned task corresponds to a human language description. Because agents can only access previously learned skills through these descriptions, the agent can always provide a human-interpretable description of its choices.
In order to help the agent learn the complex temporal dependencies necessary for the hierarchical policy, we provide it with a stochastic temporal grammar that modulates when to rely on previously learned skills and when to execute new skills.
We validate our approach on Minecraft games designed to explicitly test the ability to reuse previously learned skills while simultaneously learning new skills.

\end{abstract}


\section{Introduction}

 
Deep reinforcement learning has demonstrated success in policy search for tasks in domains like game playing \citep{Mnih2015, Silver2016, Silver2017, Kempka2016, Mirowski2017} and robotic control \citep{Levine2016, Levine2016ISER, Pinto2016}. However, it is very difficult to accumulate multiple skills using just one policy network \citet{Teh2017}. Knowledge transfer techniques like distillation \citep{Bengio2012, Rusu2016, Parisotto2016, Teh2017} have been applied to train a policy network both to learn new skills while preserving previously learned skill as well as to combine single-task policies into a multi-task policy. Existing approaches usually treat all tasks independently. This often prevents full exploration of the underlying relations between different tasks. They also typically assume that all policies share the same state space and action space. This precludes transfer of previously learned simple skills to a new policy defined over a space with differing states or actions.

When humans learn new skills, we often take advantage of our existing skills and build new capacities by composing or combining simpler ones. For instance, learning multi-digit multiplication relies on the knowledge of single-digit multiplication; learning how to properly prepare individual ingredients facilitates cooking dishes based on complex recipes.

Inspired by this observation, we propose a hierarchical policy network which can reuse previously learned skills alongside and as subcomponents of new skills. It achieves this by discovering the underlying relations between skills.

\begin{figure}[t!]
\centering
\includegraphics[trim={0 0 0 0},clip,width=0.85\linewidth]{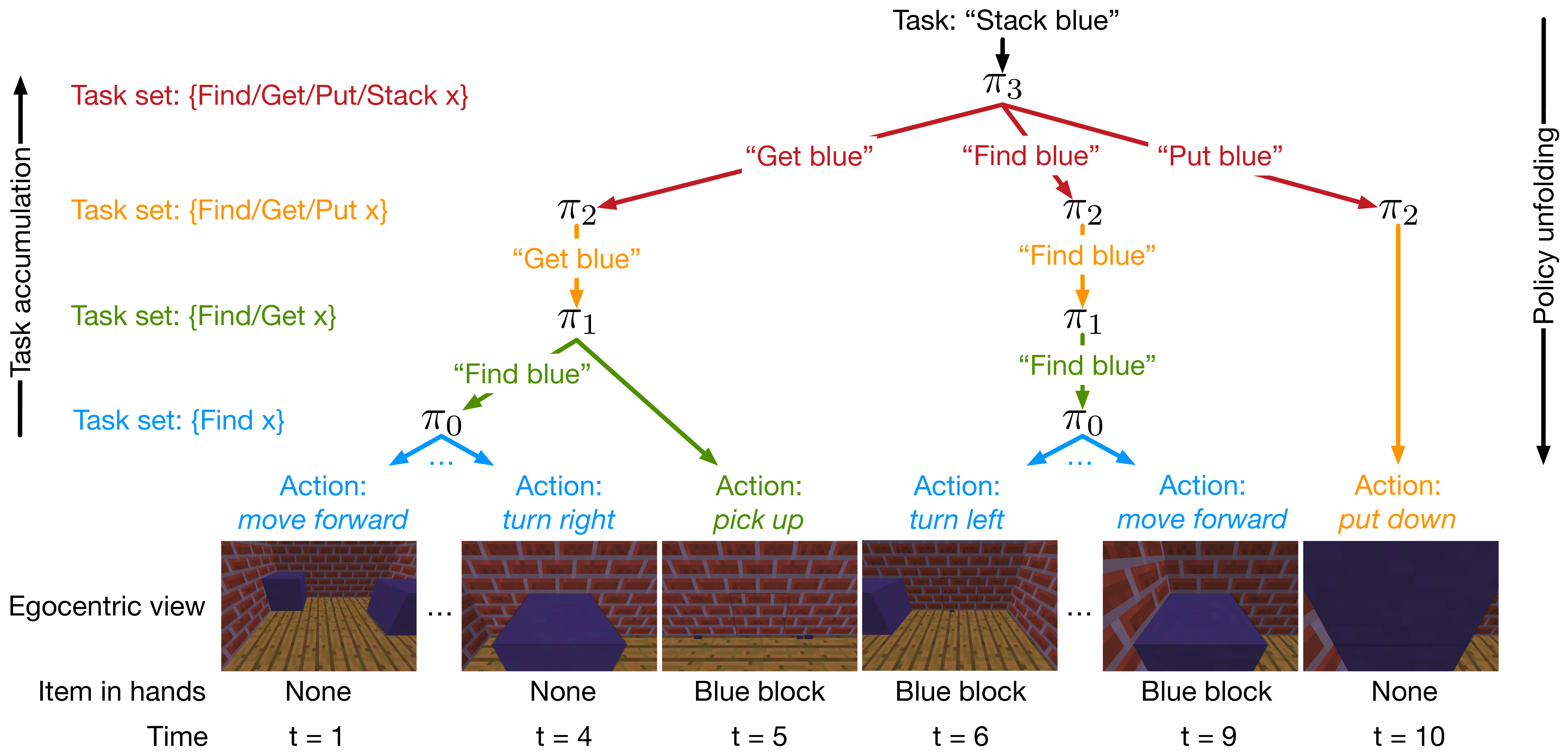}
\caption{Example of our multi-level hierarchical policy for a given task -- stacking two blue blocks. Each arrow represents one step generated by a certain policy and the colors of arrows indicate the source policies. Note that at each step, a policy either utters an instruction for the lower-level policy or directly takes an action.}
\label{fig:intro}
\end{figure}

To represent the skills and their relations in an interpretable way, we also encode all tasks using human instructions such as ``put down.'' This allows the agent to communicate its policy and generate plans using human language. Figure~\ref{fig:intro} illustrates an example: given the instruction ``Stack blue,'' our hierarchical policy learns to compose instructions and take multiple actions through a multi-level hierarchy in order to stack two blue blocks together. Steps from the top-level policy $\pi_3$ (i.e., the red branches) outline a learned high-level plan -- ``Get blue $\rightarrow$ Find blue $\rightarrow$ Put blue.'' In addition, from lower level policies, we may also clearly see composed plans for other tasks. Based on policy $\pi_2$, for instance, the task ``Get blue'' has two steps -- ``Find blue $\rightarrow$ action: turn left,'' whereas ``Put blue'' can be executed by a single action ``put down'' according to $\pi_3$. Through this hierarchical model, we may i) accumulate tasks progressively from a terminal policy to a top-level policy and ii) unfold the global policy from top-level to basic actions.

 
In order to better track temporal relationships between tasks, we train a stochastic temporal grammar (STG) model on the sequence of policy selections (previously learned skill or new skill) for positive episodes. The STG focuses on modeling priorities of tasks: for example, it is necessary to obtain an object before putting it down. Integrating the STG into the hierarchical policy boosts efficiency and accuracy by explicitly modeling such commonsense world knowledge.


We validated our approach by testing it on object manipulation tasks implemented in a Minecraft world. Our experimental results demonstrate that this framework can (i) efficiently learn hierarchical policies and representations for multi-task RL; 
(ii) learn to utter human instructions to deploy pretrained policies, improve their explainability and reuse skills; and 
(iii) learn a stochastic temporal grammar via self-supervision to predict future actions.

\section{Related Work}
\textbf{Multi-task Reinforcement Learning}. Previous work on multi-task reinforcement learning mainly falls into two families: knowledge transfer through distillation \citep{Rusu2016, Parisotto2016, Teh2017, Tessler2017} or modular policy design through 2-layer hierarchical policy \citep{Andreas2017}. 
Our multi-level policy is more similar to the latter approach. The main differences between our model and the one in \citet{Andreas2017} are two-fold: i) we do not assume that a global task can be executed by only performing predefined sub-tasks; ii) in our multi-level policy, global tasks at a lower-level layer may also be used as sub-tasks by global tasks carried out at higher-levels.

\textbf{Hierarchical Reinforcement Learning}. Complex policies often require the modeling of longer temporal dependencies than what standard Markov decision processes (MDPs) can capture. 
To combat this, hierarchical reinforcement learning was introduced to extend MDPs to semi-MDPs \citep{Sutton1999}, where options (or macro actions) are introduced on top of primitive actions to decompose the goal of a task into multiple subgoals. 
In hierarchical RL, two sets of policies are trained: local policies that map states to primitive actions for achieving subgoals, and a global policy that initiates suitable subgoals in a sequence to achieve the final goal of a task \citep{Bacon2015, Kulkarni2016, Vezhnevets2016, Tessler2017, Andreas2017}. 
%
This two-layer hierarchical policy design significantly improves the ability of discovering complex policies which  can not be learned by flat policies. However, it also makes some strict assumptions that limit its flexibility: i) a task's global policy cannot use a simpler task's policy as part of its base policies; ii) a global policy is assumed 
to be executable by only using local policies over specific options. 
In this work, we aim to learn a multi-level global policy which does not have these two assumptions. In addition, previous work usually use a latent variable to represent a task. In our work, we encode a task by a human instruction to learning task-oriented language grounding as well as to improve the interpretability of plans composed by our hierarchical policies.

\textbf{Language grounding via reinforcement learning}. Recently, there has been work on grounding human language in 3D game environments \citep{Hermann2017, Chaplot2017} or in text-based games \citep{Narasimhan2015} via reinforcement learning. In these games agents are instructed to pick up an item described by a sentence. Besides visual grounding, \citet{Andreas2017} grounded instructions (not necessarily using human language) to local policies in hierarchical reinforcement learning. Our approach not only learns the language grounding for both visual knowledge and policies, but is also trained to utter human instructions as an explicit explanation of its decisions to humans. To our knowledge, this is the first model that learns to compose plans for complex tasks based on simpler ones which have human descriptions.

\section{Model}
In this section, we discuss our multi-task RL setting, hierarchical policy, stochastic temporal grammar, and how interaction of these components can achieve plan composition.

\subsection{Multitask RL Setting}
\label{sec:multitask_setting}
Let $\mathcal{G}$ be a task set, where each task $g$ is uniquely described by a human instruction. For simplicity, we assume a two-word tuple template consisting of a skill and an item for such a phrase, i.e., $\langle u_{\text{skill}}, u_{\text{item}}\rangle$. Each tuple describes an object manipulation task. In this paper, we define $g = \langle u_{\text{skill}}, u_{\text{item}}\rangle$ by default, thus tasks and instructions are treated as interchangeable concepts.

For each task, we define a Markov decision process (MDP) represented by states $s \in \mathcal{S}$ and primitive actions $a \in \mathcal{A}$. Rewards are specified for goals of different tasks, thus we use a function $R(s, g)$ to signal the reward when performing any given task $g$.

We assume that as a starting point, we have a terminal policy $\pi_0$ (as shown in Figure~\ref{fig:flat_arch}) trained for a set of basic tasks (i.e., a terminal task set $\mathcal{G}_0$). The task set is then progressively increased as the agent is instructed to do more tasks by humans at multiple stages, such that $\mathcal{G}_0 \subset \mathcal{G}_1 \subset \cdots \subset \mathcal{G}_K$, which results in life-long learning of polices from $\pi_0$ for $\mathcal{G}_0$ to $\pi_K$ for $\mathcal{G}_K$ as illustrated by the ``task accumulation'' direction in Figure~\ref{fig:intro}. At stage $k > 0$, $\mathcal{G}_{k-1}$ is defined as the base task set of $\mathcal{G}_{k}$. The tasks in $\mathcal{G}_{k-1}$ are named as base tasks at this stage and $\pi_{k-1}$ becomes the base policy of $\pi_{k}$. Here, we utilize weak supervision from humans to define what tasks shall be augmented to the previous task set at each new stage. But in general, our model is suitable for arbitrary order of task augmentation.



\subsection{Hierarchical Policy}
\label{sec:policy}


\begin{figure}[t!]
		\centering
        \begin{subfigure}[b]{0.48\textwidth}
        		\centering
                \includegraphics[trim={0 0 0 0},clip,width=0.5\linewidth]{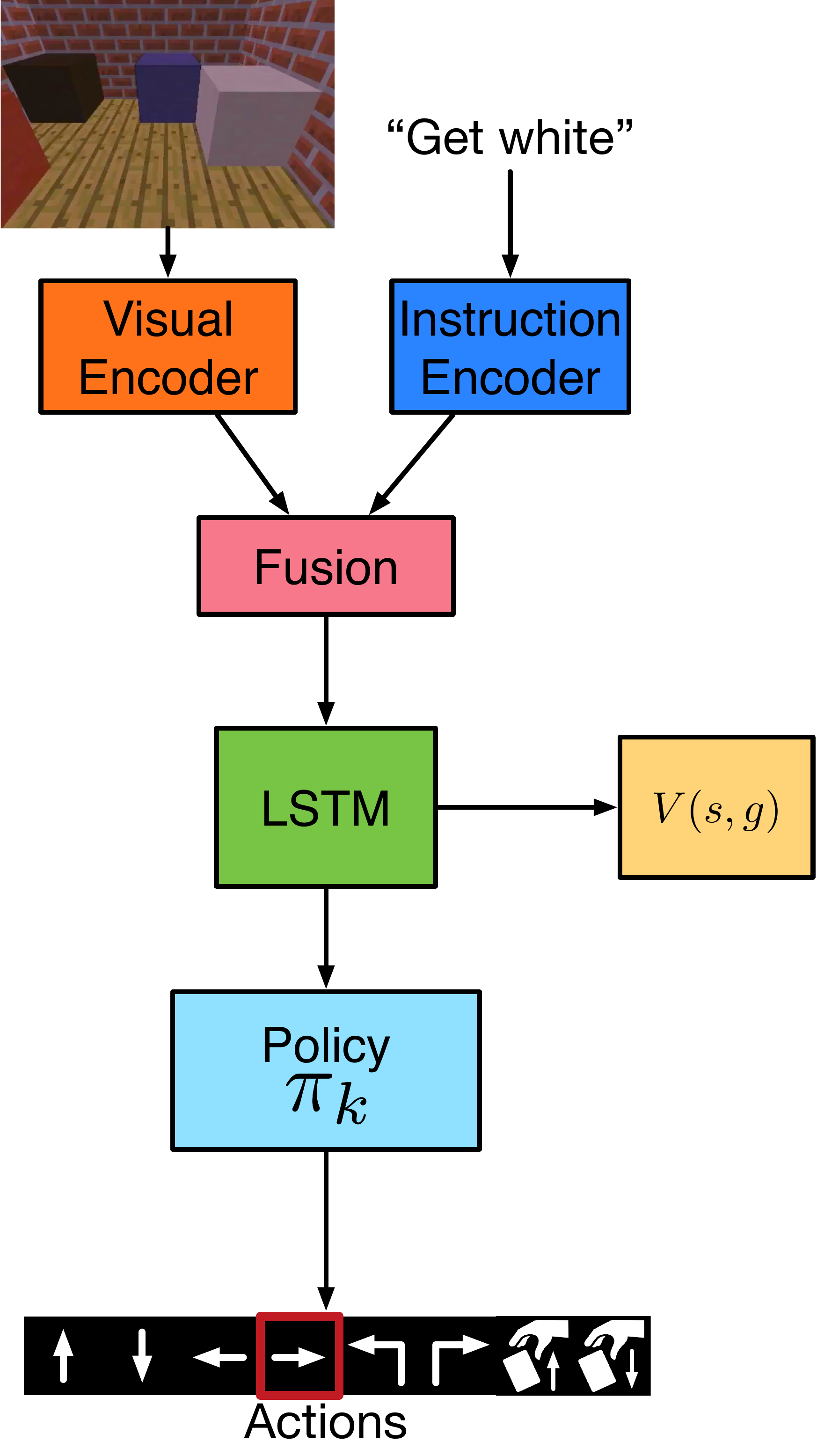}
                \caption{Flat policy.}
                \label{fig:flat_arch}
        \end{subfigure}%
        \begin{subfigure}[b]{0.48\textwidth}
        		\centering
                \includegraphics[trim={0 0 0 0},clip,width=0.9\linewidth]{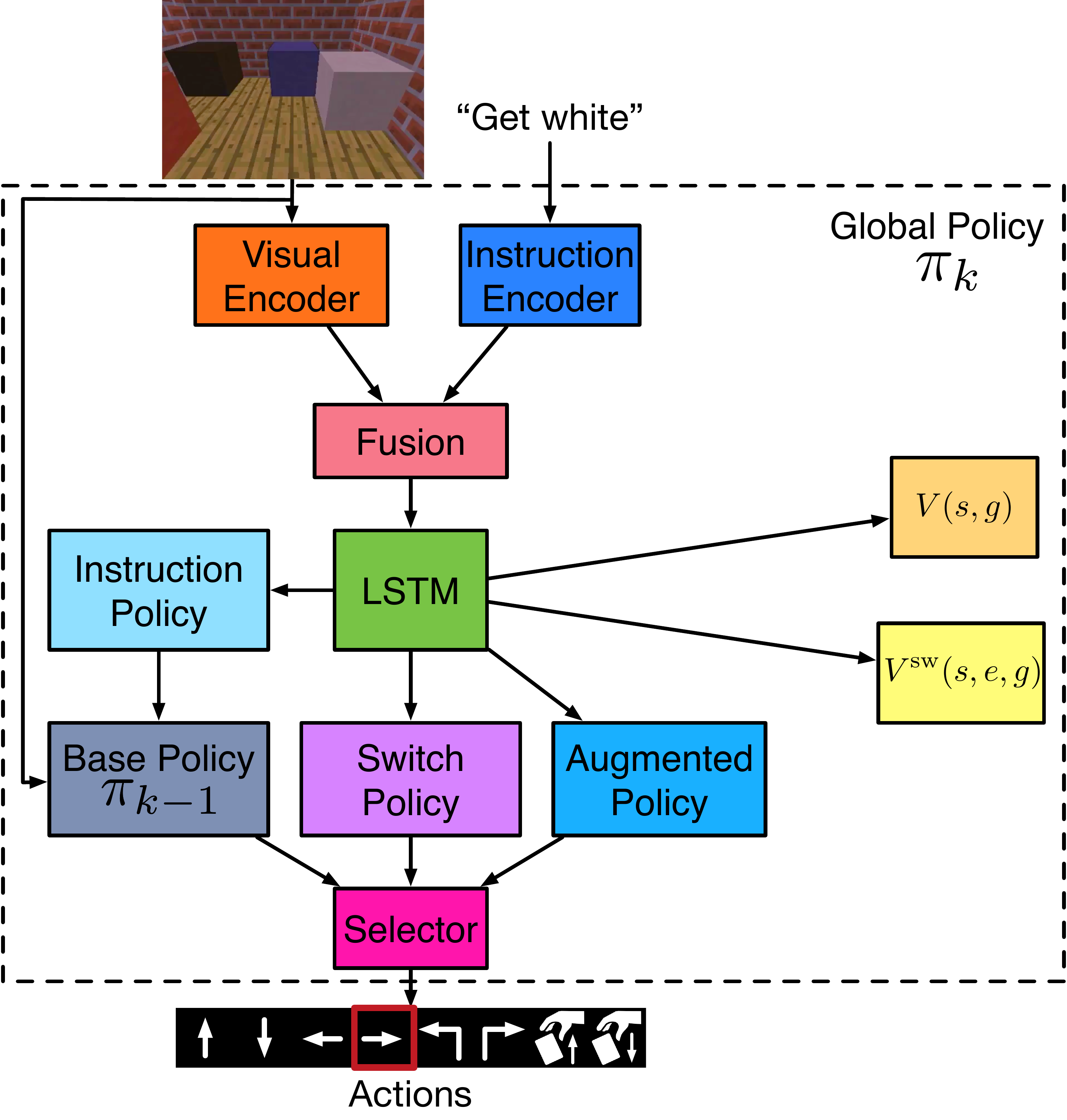}
                \caption{Hierarchical design of global policy.}
                \label{fig:HRL_arch}
        \end{subfigure}
        \caption{Flat and hierarchical policy architectures.}
        \label{fig:archs}
\end{figure}

One of our key ideas is that a new task in current task set $\mathcal{G}_k$ may be decomposed into several simpler subtasks, some of which can be base tasks in $\mathcal{G}_{k-1}$ executable by base policy $\pi_{k-1}$. Therefore, instead of using a flat policy (Figure~\ref{fig:flat_arch}) as $\pi_0$ that directly maps state and human instruction to a primitive action, we propose a hierarchical design (Figure~\ref{fig:HRL_arch}) with the ability to reuse the base policy (i.e., $\pi_{k-1}$) for performing base tasks as subtasks. Namely, at stage $k$, the global policy $\pi_k$ is defined by a hierarchical policy. This hierarchy consists of four sub-policies: a base policy for executing previously learned tasks, an instruction policy that manages communication between the global policy and the base policy, an augmented flat policy which allows the global policy to directly execute actions, and a switch policy that decides whether the global policy will primarily rely on the base policy or the augmented flat policy. 

The base policy is defined to be the global policy at the previous stage $k-1$. The instruction policy maps state $s$ and task $g \in \mathcal{G}_k$ to a base task $g^\prime \in \mathcal{G}_{k-1}$. The purpose of this policy is to inform base policy $\pi_{k-1}$ which base tasks it needs to execute. Since an instruction is represented by two words, we define the instruction policy using two conditionally independent distributions, i.e., $\pi_{k}^{\text{inst}}(g^\prime = \langle u_{\text{skill}}, u_{\text{item}}\rangle | s, g) = p_{k}^{\text{skill}}(u_{\text{skill}} | s, g) p_{k}^{\text{item}}(u_{\text{item}} | s, g)$. An augmented flat policy, $\pi_{k}^{\text{aug}}(a | s, g)$, maps state $s$ and task $g$ to a primitive action $a$ for ensuring that the global policy is able to perform novel tasks in $\mathcal{G}_{k}$ that can not be achieved by only reusing the base policy. To determine whether to perform a base task or directly perform a primitive action at each step, the global policy further includes a switch policy, $\pi_{k}^{\text{sw}}(e | s, g)$, where $e$ is a binary variable indicating the selection of the branches, $\pi_{k}^{\text{inst}}$ ($e = 0$) or $\pi_{k}^{\text{aug}}$ ($e = 1$).

Note that the above description of the hierarchical policy does not account for an STG. The instruction policy and switch policy introduced here are simplified from the ones in the full model (see Section~\ref{sec:grammar}).




At each time step, we first sample $e_t$ from our switch policy $\pi_{k}^{\text{sw}}$ to decide whether the global policy $\pi_{k}$ will rely on the base policy $\pi_{k-1}$ or the augmented flat policy $\pi_{k}^{\text{aug}}$. We also sample a new instruction $g^\prime_t$ from our instruction policy $\pi_{k}^{\text{inst}}$ in order to sample actions from the base policy. This can be summarized as:
\begin{equation}
e_t \sim \pi^\text{sw}_k(e_t | s_t, g),
\label{eq:sample_sw}
\end{equation}
\begin{equation}
g^\prime_t \sim \pi^\text{inst}_k(g^\prime_t | s_t, g),
\label{eq:sample_inst}
\end{equation}
and finally
\begin{equation}
a_t \sim \pi_k(a_t | s_t, g) = \pi_{k-1}(a_t | s_t, g^\prime_t)^{(1 - e_t)} \pi_k^\text{aug}(a_t | s_t, g)^{e_t},
\label{eq:global}
\end{equation}
where $\pi_k$ and $\pi_{k-1}$ are the global policies at stage $k$ and $k-1$ respectively. After each step, we will also obtain a reward $r_t = R(s_t, g)$.

\subsection{Stochastic Temporal Grammar}\label{sec:grammar}



Different tasks may have temporal relations. For instance, to move an object, one needs to first find and pick up that object. There has been previous research \citep{Si2011, Pirsiavash2014} using stochastic grammar models to capture such temporal relations. 
Inspired by this, we summarize temporal transitions between various tasks with an stochastic temporal grammar (STG). In our full model, the STG interacts with the hierarchical policy described above through modified switch policy and instruction policy by using the STG as a prior. This amounts to treating the past history of switches and instructions in positive episodes as a guidance on whether the hierarchical policy should defer to the base policy to execute a specific base task or employ its own augmented flat policy to take a primitive action. 

In an episode, the temporal sequence of $e_t$ and $g^\prime_t$, i.e., $\{\langle e_t, g^\prime_t \rangle; t \geq 0 \}$, can be seen as a finite state Markov chain \citep{Baum1966}. Note that the state here is referred to the tuple $\langle e_t, g^\prime_t \rangle$, which is not the state of the game $s_t \in \mathcal{S}$ defined in Section~\ref{sec:multitask_setting}. Consequently, at each level $k > 0$, we may define an STG of a task $g$ by i) transition probabilities, $\rho_k(e_t, g^\prime_t | e_{t-1}, g^\prime_{t-1}, g)$, and ii) the distribution of $\langle e_0, g^\prime_0 \rangle$, $q_k(e_0, g^\prime_0 | g)$.

With the estimated probabilities, we sample $e_t$ and $g^\prime_t$ in an episode at level $k > 0$ w.r.t. to reshaped policies ${\pi_k^{{sw}^\prime}}$ and ${\pi_k^\text{inst}}^\prime$ respectively:
\begin{itemize}
\item If $t = 0$,
\begin{equation}
e_0 \sim {\pi_k^\text{sw}}^\prime(e_0 | s_t, g) \propto \pi^\text{sw}_k(e_0 | s_t, g)\sum_{g^\prime \in \mathcal{G}_{k-1}}q_k(e_0, g^\prime | g),
\end{equation}
\begin{equation}
g^\prime_0 \sim {\pi_k^\text{inst}}^\prime(g^\prime_0 | s_t, g) \propto \pi_k^\text{inst}(g^\prime_0 | s_t, g)q_k(e_0 = 0, g^\prime_0 | g); 
\end{equation}
\item Otherwise,
\begin{equation}
e_t \sim {\pi^\text{sw}_k}^\prime(e_t | e_{t-1}, g^\prime_{t-1}, s_t, g) \propto \pi^\text{sw}_k(e_t | s_t, g)\sum_{g^\prime \in \mathcal{G}_{k-1}}\rho_k(e_t, g^\prime | e_{t-1}, g^\prime_{t-1}, g), 
\end{equation}
\begin{equation}
g^\prime_t \sim {\pi_k^\text{inst}}^\prime(g^\prime_t | e_{t-1}, g^\prime_{t-1}, s_t, g) \propto \pi_k^\text{inst}(g^\prime_t | s_t, g)\rho_k(e_t = 0, g^\prime_t | e_{t-1}, g^\prime_{t-1}, g).
\end{equation}
\end{itemize}
Note that primitive action sampling is not affected by the STG.


\subsection{Plan Composition}

Combined with our hierarchical policy and STG defined above, we are able to run an episode to compose a plan for a task specified by a human instruction. Algorithm~\ref{alg:rollout} in Appendix~\ref{sec:app_algorithms} summarized this procedure with respect to the policy and STG at level $k$. Note that to fully utilize the base policy, we assume that once triggered, a base policy will play to the end before the global policy considers the next move.

\section{Learning}
\label{sec:learning}
The learning algorithm is outlined in Algorithm~\ref{alg:learning} in Appendix~\ref{sec:app_algorithms}. We learn our final hierarchical policy through $k$ stages of skill acquisition. Each of these stages is broken down into a base skill acquisition phase and a novel skill acquisition phase in a 2-phase curriculum learning.

In the base skill acquisition phase, we only sample tasks from the base task set $\mathcal{G}_{k-1}$. This ensures that the global policy learns how to use previously learned skills by issuing instructions to the base policy. In other words, this phase teaches the agent how to connect its instruction policy to its base policy. Once the average reward for all base tasks exceeds a certain threshold, we proceed to the next phase.

In the novel skill acquisition phase, we sample tasks from the full task set, $\mathcal{G}_k$, for the $k$-th stage of skill acquisition. It is in this phase that the agent can learn when to rely on the base policy and when to rely on the augmented flat policy for executing novel tasks.

In each of these phases, all policies are trained with advantage actor-critic (A2C) (Section~\ref{sec:gradient}) and distributions in the STG are estimated based on accumulated positive episodes (Section~\ref{sec:learning_STGs}). 

\subsection{Policy Optimization by Advantage Actor-Critic}
\label{sec:gradient}

We use advantage actor-critic (A2C) for policy optimization with off-policy learning \citep{Su2017}. Here, we only consider the gradient for global policies (i.e., $k > 0$) as we assume the terminal policy has been trained as initial condition. 
Let $V_k(s_t, g)$ be a value function indicating the expected return given state $s_t$ and task $g$. To reflect the nature of the branch switching in our model, we introduce another value function $V_k^\text{sw}(s_t, e_t, g)$ to represent the expected return given state $s_t$, task $g$ and current branch selection $e_t$. 

Thus, given a trajectory $\Gamma = \{\langle s_t, e_t, g^\prime_t, a_t, r_t, \mu^\text{sw}_k(\cdot | s_t), \mu_k^\text{inst}(\cdot | s_t, g), \mu_{k}^\text{aug}(\cdot | s_t, g), g \rangle: t = 0, 1, \cdots, T\}$ generated by old policies $\mu^\text{sw}_k(\cdot | s_t)$, $\mu_k^\text{inst}(\cdot | s_t, g)$, and $\mu_{k}^\text{aug}(\cdot | s_t, g)$, the policy gradient reweighted by importance sampling can be formulated as
\begin{equation}
\begin{array}{ll}
&\underbrace{\omega_t^\text{sw}\nabla_{\theta^\text{sw}} \log\pi_k^\text{sw}(e_t | s_t, g)A(s_t, g, e_t)}_{\text{1st term: switch policy gradient}}\\
+ & \underbrace{(1 - e_t)\omega_t^\text{inst}\nabla_{\theta^\text{inst}} \log\pi_k^\text{inst}(g^\prime_t | s_t, g)A(s_t, g, e_t, g^\prime_t)}_{\text{2nd term: instruction policy gradient}} \\
+ & \underbrace{e_t\omega_t^\text{aug}\nabla_{\theta^\text{aug}} \log\pi_k^\text{aug}(a_t | s_t, g)A(s_t, g, e_t, a_t)}_{\text{3rd term: augmented policy gradient}},
\end{array}
\label{eq:policy_gradient}
\end{equation}
where $\omega_t^\text{sw} = \frac{\pi_k^\text{sw}(e_t | s_t, g)}{\mu_k^\text{sw}(e_t | s_t, g)}$, $\omega_t^\text{inst} = \frac{\pi_k^\text{inst}(g^\prime_t | s_t, g)}{\mu_k^\text{inst}(g^\prime_t | s_t, g)}$, and $\omega_t^\text{aug} = \frac{\pi_k^\text{aug}(a_t | s_t, g)}{\mu_k^\text{aug}(a_t | s_t, g)}$ are importance sampling weights for the three terms respectively; $A(s_t, g, e_t)$, $A(s_t, g, e_t, g^\prime_t)$, and $A(s_t, g, e_t, a_t)$ are estimates of advantage functions, which have multiple possible definitions. In this paper, we define them by the difference between empirical return and value function estimation: $A(s_t, g, e_t) = \sum_{\tau=0}^\infty \gamma^\tau R(s_{t+\tau}, g) - V_k(s_t, g)$, $A(s_t, g, e_t, g^\prime_t) = A(s_t, g, e_t, a_t) = \sum_{\tau=0}^\infty \gamma^\tau R(s_{t+\tau}, g) - V_k^\text{sw}(s_t, g, e_t)$, where $\gamma$ is the discounted coefficient.

Finally, the value functions can be updated using the following gradient:
\begin{equation}
\nabla_{\theta_{v}}\frac{1}{2} \left[\sum_{\tau=0}^\infty \gamma^\tau R(s_{t+\tau}, g) - V_k(s_t, g)\right]^2 + \nabla_{\theta_{v}^\text{sw}}\frac{1}{2} \left[\sum_{\tau=0}^\infty \gamma^\tau R(s_{t+\tau}, g) - V_k^\text{sw}(s_t, e_t, g)\right]^2.
\label{eq:value_gradient}
\end{equation}

To increase the episode efficiency, after running an episode, we conduct $n$ mini-batch updates where $n$ is sampled from a Poisson distribution with $\lambda = 4$, similar to \citet{Wang2017}. Note that one can also apply other common policy optimization methods, e.g., A3C \citep{Mnih2016}, to our model. We leave this as future work to evaluate the efficiency of different methods when using our model.

Optimizing all three sub-policies together leads to unstable learning. To avoid this, we apply a simple alternating update procedure. For each set of $M$ iterations, we keep two of the sub-policies fixed and train only the single policy that remains. When we reach $M$ iterations, we switch the policy that is trained. For all experiments in this paper, we use $M=500$. This alternating update procedure is used within both phases of curriculum learning. 


\subsection{Learning an STG}
\label{sec:learning_STGs}
If at any point in the aforementioned training process the agent receives a positive reward after an episode, we update the stochastic temporal grammar. $\rho_k$ and $q_k$ of the STG are both initialized to be uniform distributions. Since the STG is a finite state Markov chain over tuples $\langle e_t, g^\prime_t \rangle$, we use maximum likelihood estimation (MLE) to update the distributions \citep{Baum1966}. As the training progresses, the STG starts to guide the exploration.


To avoid falling into local minima in the early stages of training, it is important to encourage random exploration in early episodes. Based on our experiments, we find that using $\epsilon$-greedy suffices.



\section{Experiments}
\subsection{Game Environment and Task Specifications}

\begin{wrapfigure}{R}{0.15\textwidth}
\centering
\includegraphics[trim={0 0 0 0},clip,width=1.0\linewidth]{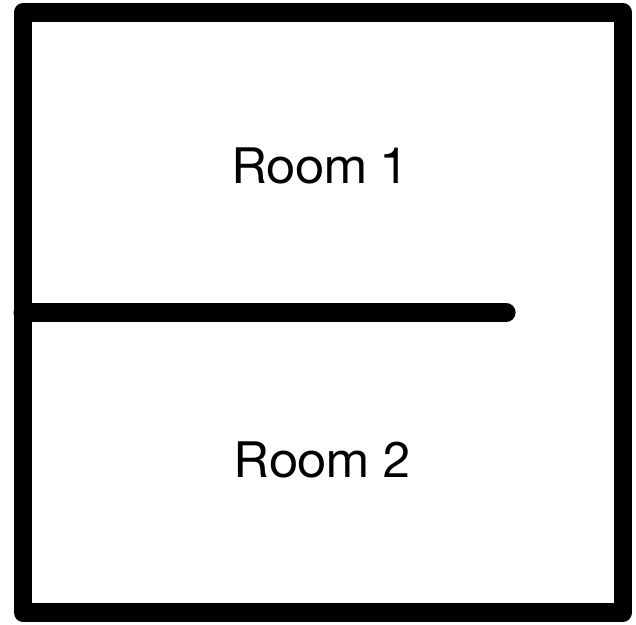}
\caption{Room layout.}\label{fig:room_layout}
\vspace{-10pt}
\end{wrapfigure}


Figure~\ref{fig:room_layout} shows the two room environment in Minecraft that we created using the Malmo platform \citep{Johnson2016}. In each episode, an arbitrary number of blocks with different colors (totally 6 colors in our experiments) are randomly placed in one of the two rooms. The agent is initially placed in the same room with the items. We consider four sets of tasks: i) $\mathcal{G}^{(0)} = \{\text{``Find x''}\}$, walking to the front of a block with color x, ii) $\mathcal{G}^{(1)} = \{\text{``Get x''}\}$, picking up a block with color x, iii) $\mathcal{G}^{(2)} = \{\text{``Put x''}\}$, putting down a block with color x, and iv) $\mathcal{G}^{(3)} = \{\text{``Stack x''}\}$, stacking two blocks with color x together. In total, there are 24 tasks. An agent can perform the following actions: ``move forward,'' ``move backward,'' ``move left,'' ``move right,'' ``turn left,'' ``turn right,'' ``pick up,'' ``put down.'' 

Without loss of generality, we assume the following skill acquisition order: $\mathcal{G}_k = \cup_{\kappa = 1}^k \mathcal{G}^{(\kappa)}$, $\forall k = 0,1,2,3$, which is a natural way to increase skill sets. One may also alter the order, and the main conclusions shall still hold. This results in policies $\{\pi_k: k = 0, 1, 2, 3\}$ for these four task sets.

We adopt a sparse reward function: when reaching the goal of a task, the agent gets a $+1$ reward; when generating an instruction $g^\prime$ that is not executable in current game (e.g., trying to find an object that does not exist in the environment), we give a $-0.5$ reward; otherwise, no reward will be given. Whenever a non-zero reward is given, the game terminates.

\subsection{Implementation Details}


We specify the architecture of the modules in our model in Appendix~\ref{sec:app_details}, where the visual and instruction encoding modules have the same architectures as the ones in \citet{Hermann2017}. We train the network with RMSProp \citep{Tieleman2012} with a learning rate of 0.0001. We set the batch size to be 36 and clip the gradient to a unit norm. For all tasks, the discounted coefficient is $\gamma = 0.95$. For the 2-phase curriculum learning, we set the average reward threshold to be 0.9 (average rewards are estimated from the most recent 200 episodes of each task).

To encourage random exploration, we apply $\epsilon$-greedy to the decision sampling for the global policy (i.e., only at the top level $k$ at each stage $k > 0$), where $\epsilon$ gradually decreases from $0.1$ to $0$. 

\begin{figure}[h!]
		\centering
        \begin{subfigure}[b]{0.35\textwidth}
        		\centering
                \includegraphics[trim={10 0 50 10},clip,width=1.0\linewidth]{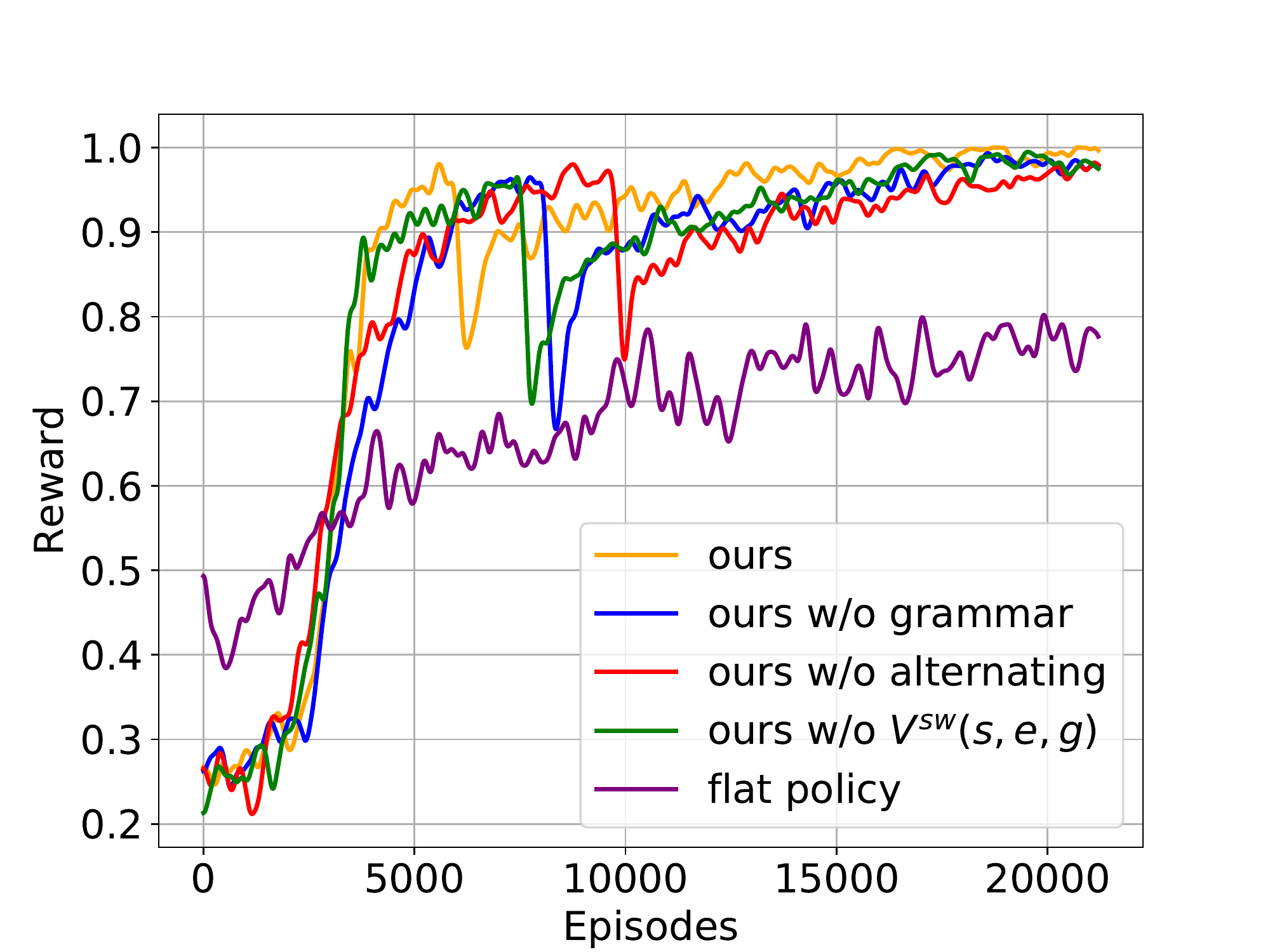}
                \caption{Learning curves for $\pi_1$.}
                \label{fig:get}
        \end{subfigure}%
        \begin{subfigure}[b]{0.35\textwidth}
        		\centering
                \includegraphics[trim={10 0 50 10},clip,width=1.0\linewidth]{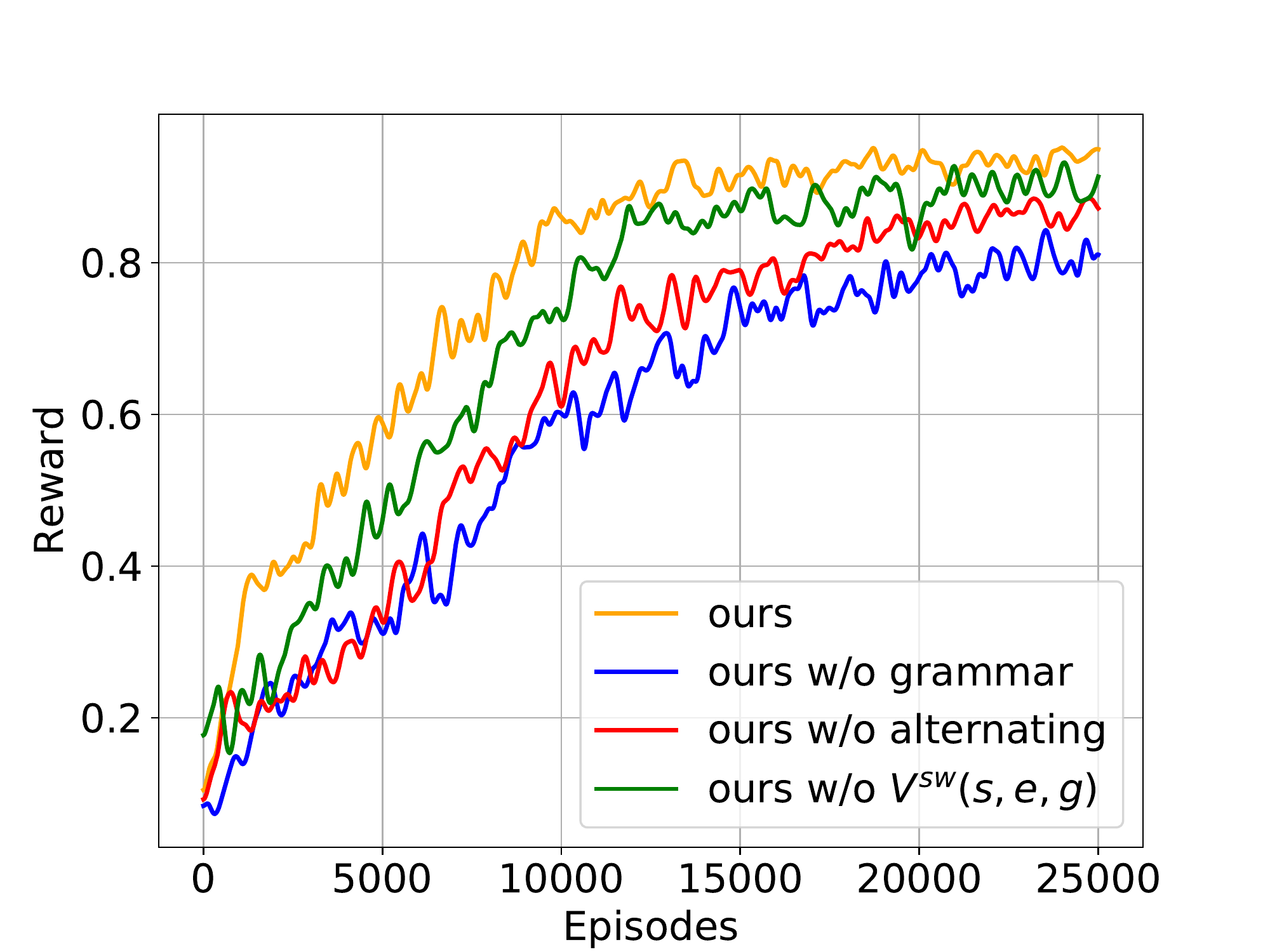}
                \caption{Learning phase 2 for $\pi_3$.}
                \label{fig:stack_phase2}
        \end{subfigure}
        \caption{Comparison of learning efficiency on two task sets: (a) $\mathcal{G}_1$ for global policy $\pi_1$ and (b) $\mathcal{G}_3$ for global policy $\pi_3$ respectively.}\label{fig:rewards}
\end{figure}

\begin{figure}[h!]
		\centering
        \begin{subfigure}[b]{0.3\textwidth}
        		\centering
                \includegraphics[trim={10 0 50 10},clip,width=1.0\linewidth]{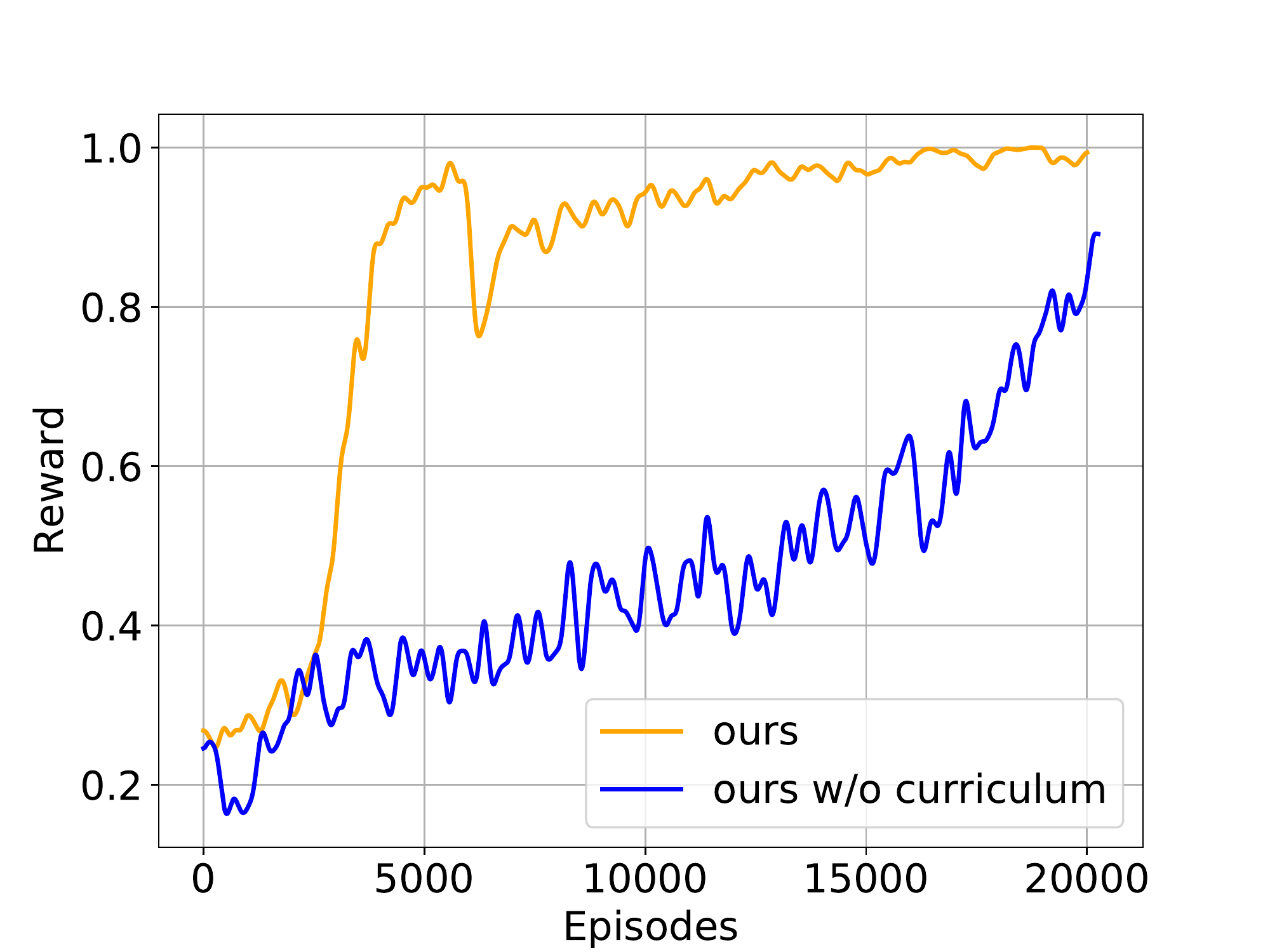}
                \caption{Learning curves for $\pi_1$.}
                \label{fig:get_full}
        \end{subfigure}%
        \begin{subfigure}[b]{0.3\textwidth}
        		\centering
                \includegraphics[trim={10 0 50 10},clip,width=1.0\linewidth]{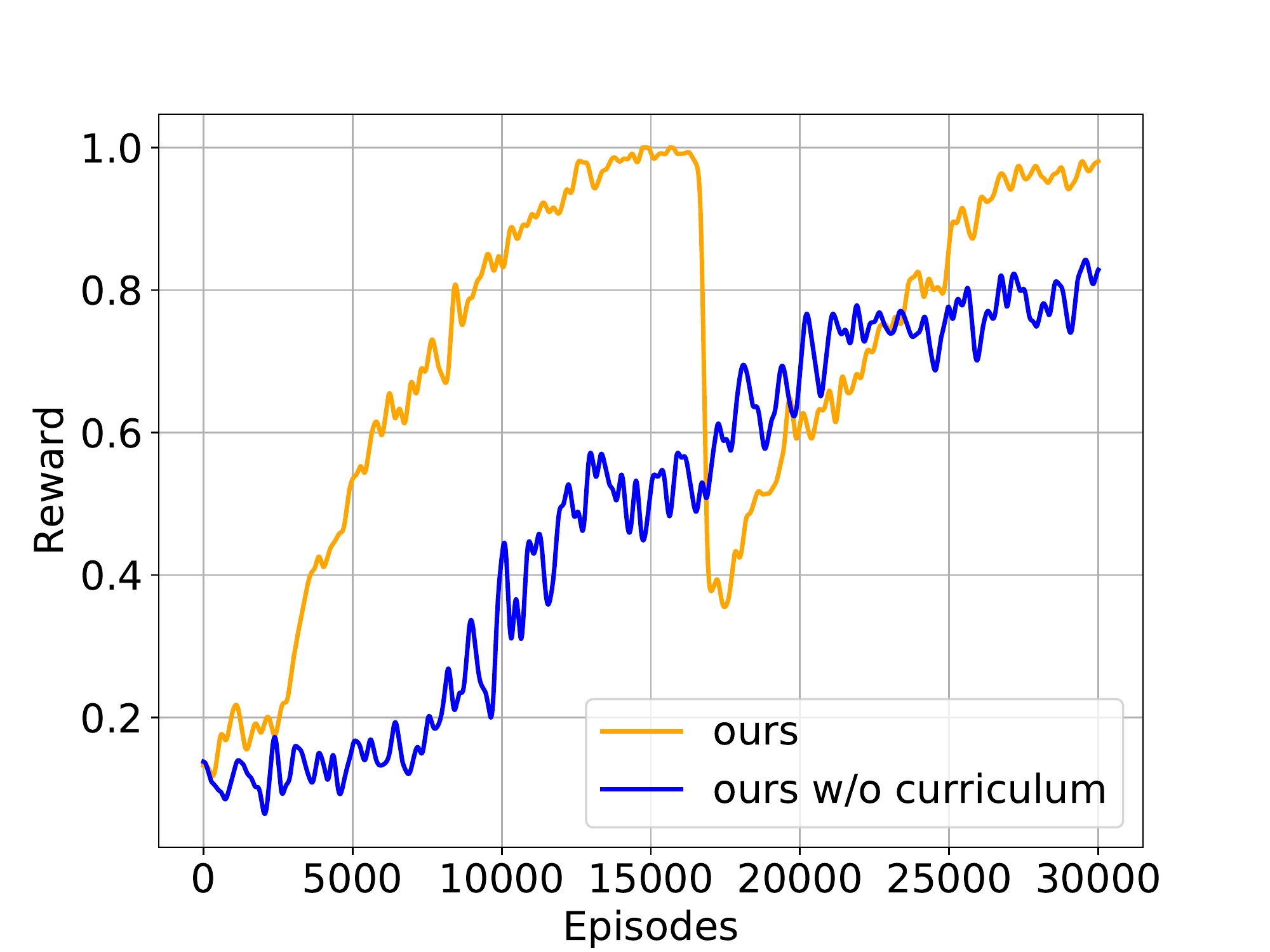}
                \caption{Learning curves for $\pi_3$.}
                \label{fig:stack_full}
        \end{subfigure}
        \caption{Effect of our 2-phase curriculum learning.}\label{fig:rewards_curr}
\end{figure}

\subsection{Learning Efficiency}

To evaluate the learning efficiency, we compare our full model with 1) a flat policy (Figure~\ref{fig:flat_arch}) as in \citet{Hermann2017} fine-tuned on the terminal policy $\pi_0$ and variants of our approach: 2) ours without STG, 3) ours without alternating policy optimization, and 4) ours without $V_k^\text{sw}(s, e, g)$ (replaced by $V_k(s, g)$ instead). Note that all the rewards have been converted to the same range, i.e., $[0, 1]$ for the sake of fair comparison. 

In Figure~\ref{fig:get}, we use various methods to train policy $\pi_1$ for the task set $\mathcal{G}_1$ based on the same base policy $\pi_0$. The large dip in the reward indicates that the curriculum learning switches from phase 1 to phase 2. From Figure~\ref{fig:get}, we may clearly see that our full model and variants can all converge within 22,000 episodes, whereas the average reward of the flat policy is still below 0.8 given the same amount of episodes. In addition, our full model finishes phase 1 significantly faster than other methods and its curve of average reward maintains notably higher than the remaining ones.

To further examine the learning efficiency during phase 2 when new tasks are added into the training process, we first pretrain $\pi_3$ 
using our full model following our definition of phase 1 in the curriculum learning. We then proceed to learning phase 2 using different approaches all based on this pretrained policy. As shown in Figure~\ref{fig:stack_phase2}, our full model has the fastest convergence and the highest average reward upon convergence. By comparing Figure~\ref{fig:get} and Figure~\ref{fig:stack_phase2}, we further show that our full model has a bigger advantage when learning more complex tasks.


To demonstrate the effect of our 2-phase curriculum learning on the training efficiency, we visualize the learning curves of our model trained with and without the curriculum learning respectively in Figure~\ref{fig:rewards_curr}. According to the results, the curriculum learning indeed helps accelerate the convergence, which empirically proves the importance of encouraging a global policy to reuse relevant skills learned by its base policy.


\begin{table}[t]
\caption{Success rates in different game environments. All policies are trained in a small room.}
\begin{center}
\begin{tabular}{ c|cccc|cccc } 
\multirow{2}{*}{\bf Method} & \multicolumn{4}{c|}{\bf Small room} & \multicolumn{4}{c}{\bf Big room} \\ [3pt] \cline{2-9}
& {\bf Find x} & {\bf Get x} & {\bf Put x} & {\bf Stack x} & {\bf Find x} & {\bf Get x} & {\bf Put x} & {\bf Stack x} \\[3pt] 
 Full model  & 0.995 & 0.970 & 1.00 & 0.955 & 0.723 & 0.648 & 1.00 & 0.613 \\  [3pt]
 Flat policy & 0.980 & 0.965 & 1.00 & - & 0.515 & 0.450 & 1.00 & - \\ [3pt]
\end{tabular}
\end{center}
\label{tab:gen}
\end{table}

\subsection{Policy Generalization}
Finally, we evaluate how the hierarchical design and encoding tasks by human instructions benefit the generalization of learned policies in the following two ways.

First, we train $\pi_1$ in a simpler setting where in each episode, only one item (i.e, the target item of the given task) is present. We then test the policy $\pi_1$ for ``Get x'' tasks in a room where there will be multiple items serving as distraction and the agent must interact with the correct one. Both the flat policy and the hierarchical policy can achieve near perfect testing success rate in the simple setting. However, in the more complex setting, flat policy can not differentiate the target item from other items that are also placed in the room (the success rate drops to 29\%), whereas our hierarchical policy still maintains a high success rate (94\%). This finding suggests that the hierarchical policy not only picks up the concept of ``find'' and ``get'' skills as the flat policy does, but also inherits the concept of items from the base policy by learning to utter correct instructions to deploy ``find'' skill in the base policy.

Second, we remove the wall between the two rooms shown in Figure~\ref{fig:room_layout} and test the flat policy and our full model in this bigger room for various tasks. Both policies are trained in the two small rooms. There are multiple items in a room for both training and testing cases. Note that the success rate of stacking tasks by the flat policy is not shown here since it is extremely difficult for the flat policy to converge to a decent policy for the full task set $\mathcal{G}_3$ even after 4 days of training. The success rates are summarized in Tab.~\ref{tab:gen}. Using the flat policy results in a much bigger drop in the testing success rate compared to using out full model. This is mainly because that our global policy will repeatedly call its base policy to execute the same task until the agent finally achieves the goal even though the trained agent is unable to reach the goal by just one shot due to the simplicity of the training environment.

\subsection{Policy Interpretability}

We visualize typical hierarchical plans of several tasks generated by global policies learned by our full model in Appendix \ref{sec:app_viz} (Figure~\ref{fig:viz_policies}). It can been seen from the examples that our global policies adjust the composed plans in different scenarios. For instance, in the second plan on the first row, $\pi_1$ did not deploy base policy $\pi_0$ as the agent was already in front of the target item at the beginning of the episode, whereas in the plan on the second row, $\pi_1$ deployed $\pi_0$ for the ``Find x'' base task twice consecutively, as it did not finish the base task in the first call.


\section{Conclusion}

In this work, we have proposed a hierarchal policy modulated by a stochastic temporal grammar as a novel framework for efficient multi-task reinforcement learning through multiple training stages. Each task in our settings is described by a human instruction. The resulting global policy is able to reuse previously learned skills for new tasks by generating corresponding human instructions to inform base policies to execute relevant base tasks. We evaluate this framework in Minecraft games and have shown that our full model i) has a significantly higher learning efficiency than a flat policy does, ii) generalizes well in unseen environments, and iii) is capable of composing hierarchical plans in an interpretable manner.

Currently, we rely on weak supervision from humans to define what skills to be learned in each training stage. In the future, we plan to automatically discover the optimal training procedures to increase the task set.


\bibliography{iclr2018_conference}

\begin{thebibliography}{29}
\providecommand{\natexlab}[1]{#1}
\providecommand{\url}[1]{\texttt{#1}}
\expandafter\ifx\csname urlstyle\endcsname\relax
  \providecommand{\doi}[1]{doi: #1}\else
  \providecommand{\doi}{doi: \begingroup \urlstyle{rm}\Url}\fi

\bibitem[Andreas et~al.(2017)Andreas, Klein, and Levine]{Andreas2017}
Jacob Andreas, Dan Klein, and Sergey Levine.
\newblock Modular multitask reinforcement learning with policy sketches.
\newblock In \emph{International Conference on Machine Learning (ICML)}, 2017.

\bibitem[Bacon \& Precup(2015)Bacon and Precup]{Bacon2015}
Pierre-Luc Bacon and Doina Precup.
\newblock The option-critic architecture.
\newblock In \emph{NIPS Deep Reinforcement Learning Workshop}, 2015.

\bibitem[Baum \& Petrie(1966)Baum and Petrie]{Baum1966}
Leonard~E. Baum and Ted Petrie.
\newblock Statistical inference for probabilistic functions of finite state
  markov chains.
\newblock \emph{The Annals of Mathematical Statistics}, 37\penalty0
  (6):\penalty0 1554--1563, 1966.

\bibitem[Bengio(2012)]{Bengio2012}
Yoshua Bengio.
\newblock Deep learning of representations for unsupervised and transfer
  learning.
\newblock In \emph{JMLR: Workshop on Unsupervised and Transfer Learning}, 2012.

\bibitem[Chaplot et~al.(2017)Chaplot, Sathyendra, Pasumarthi, Rajagopal, and
  Salakhutdinov]{Chaplot2017}
Devendra~Singh Chaplot, Kanthashree~Mysore Sathyendra, Rama~Kumar Pasumarthi,
  Dheeraj Rajagopal, and Ruslan Salakhutdinov.
\newblock Gated-attention architectures for task-oriented language grounding.
\newblock \emph{arXiv preprint arXiv:1706.0723}, 2017.

\bibitem[Hermann et~al.(2017)Hermann, Hill, Green, Wang, Faulkner, Soyer,
  Szepesvari, Czarnecki, Jaderberg, Teplyashin, Wainwright, Apps, Hassabis, and
  Blunsom]{Hermann2017}
Karl~Moritz Hermann, Felix Hill, Simon Green, Fumin Wang, Ryan Faulkner, Hubert
  Soyer, David Szepesvari, Wojtek Czarnecki, Max Jaderberg, Denis Teplyashin,
  Marcus Wainwright, Chris Apps, Demis Hassabis, and Phil Blunsom.
\newblock Grounded language learning in a simulated 3d world.
\newblock \emph{arXiv preprint arXiv:1706.06551}, 2017.

\bibitem[Johnson et~al.(2016)Johnson, Hofmann, Hutton, and
  Bignell]{Johnson2016}
Matthew Johnson, Katja Hofmann, Tim Hutton, and David Bignell.
\newblock The malmo platform for artificial intelligence experimentation.
\newblock In \emph{International Joint Conference on Artificial Intelligence
  (IJCAI)}, 2016.

\bibitem[Kempka et~al.(2016)Kempka, Wydmuch, Runc, Tocze, and
  Ja{\'s}kowski]{Kempka2016}
Micha{\l} Kempka, Marek Wydmuch, Grzegorz Runc, Jakub Tocze, and Wojciech
  Ja{\'s}kowski.
\newblock Vizdoom: A doom-based ai research platform for visual reinforcement
  learning.
\newblock In \emph{IEEE Conference on Computational Intelligence and Games
  (CIG)}, 2016.

\bibitem[Kulkarni et~al.(2016)Kulkarni, Narasimhan, Saeedi, , and
  Tenenbaum]{Kulkarni2016}
Tejas~D. Kulkarni, Karthik Narasimhan, Ardavan Saeedi, , and Josh Tenenbaum.
\newblock Hierarchical deep reinforcement learning: Integrating temporal
  abstraction and intrinsic motivation.
\newblock In \emph{Advances in Neural Information Processing Systems (NIPS)},
  2016.

\bibitem[Levine et~al.(2016{\natexlab{a}})Levine, Finn, Darrell, and
  Abbeel]{Levine2016}
Sergey Levine, Chelsea Finn, Trevor Darrell, and Pieter Abbeel.
\newblock End-to-end training of deep visuomotor policies.
\newblock \emph{Journal of Machine Learning Research}, 17\penalty0
  (39):\penalty0 1--40, 2016{\natexlab{a}}.

\bibitem[Levine et~al.(2016{\natexlab{b}})Levine, Pastor, Krizhevsky, Ibarz, ,
  and Quillen]{Levine2016ISER}
Sergey Levine, Peter Pastor, Alex Krizhevsky, Julian Ibarz, , and Deirdre
  Quillen.
\newblock Learning hand-eye coordination for robotic grasping with deep
  learning and large-scale data collection.
\newblock In \emph{'' International Symposium on Experimental Robotics (ISER)},
  2016{\natexlab{b}}.

\bibitem[Mirowski et~al.(2017)Mirowski, Pascanu, Viola, Soyer, Ballard, Banino,
  Denil, Goroshin, Sifre, Kavukcuoglu, Kumaran, and Hadsell]{Mirowski2017}
Piotr Mirowski, Razvan Pascanu, Fabio Viola, Hubert Soyer, Andy Ballard, Andrea
  Banino, Misha Denil, Ross Goroshin, Laurent Sifre, Koray Kavukcuoglu,
  Dharshan Kumaran, and Raia Hadsell.
\newblock Learning to navigate in complex environments.
\newblock In \emph{International Conference on Learning Representations
  (ICLR)}, 2017.

\bibitem[Mnih et~al.(2015)Mnih, Kavukcuoglu, Silver, Rusu, Veness, Bellemare,
  Graves, Riedmiller, Fidjeland, Ostrovski, Petersen, Beattie, Sadik,
  Antonoglou, King, Kumaran, Wierstra, Legg, and Hassabis]{Mnih2015}
Volodymyr Mnih, Koray Kavukcuoglu, David Silver, Andrei~A. Rusu, Joel Veness,
  Marc~G. Bellemare, Alex Graves, Martin Riedmiller, Andreas~K. Fidjeland,
  Georg Ostrovski, Stig Petersen, Charles Beattie, Amir Sadik, Ioannis
  Antonoglou, Helen King, Dharshan Kumaran, Daan Wierstra, Shane Legg, and
  Demis Hassabis.
\newblock Human-level control through deep reinforcement learning.
\newblock \emph{Nature}, 518\penalty0 (7540):\penalty0 529--533, 2015.

\bibitem[Mnih et~al.(2016)Mnih, Badia, Mirza, Graves, Lillicrap, Harley,
  Silver, and Kavukcuoglu]{Mnih2016}
Volodymyr Mnih, Adria~Puigdomenech Badia, Mehdi Mirza, Alex Graves, Timothy
  Lillicrap, Tim Harley, David Silver, and Koray Kavukcuoglu.
\newblock Asynchronous methods for deep reinforcement learning.
\newblock In \emph{International Conference on Machine Learning (ICML)}, 2016.

\bibitem[Narasimhan et~al.(2015)Narasimhan, Kulkarni, and
  Barzilay]{Narasimhan2015}
Karthik Narasimhan, Tejas Kulkarni, and Regina Barzilay.
\newblock Language understanding for text-based games using deep reinforcement
  learning.
\newblock In \emph{Proceedings of the Conference on Empirical Methods in
  Natural Language Processing (EMNLP)}, 2015.

\bibitem[Parisotto et~al.(2016)Parisotto, Ba, and Salakhutdinov]{Parisotto2016}
Emilio Parisotto, Jimmy~Lei Ba, and Ruslan Salakhutdinov.
\newblock Actor-mimic: Deep multitask and transfer reinforcement learning.
\newblock In \emph{International Conference on Learning Representations
  (ICLR)}, 2016.

\bibitem[Pinto \& Gupta(2016)Pinto and Gupta]{Pinto2016}
Lerrel Pinto and Abhinav Gupta.
\newblock Supersizing self-supervision: Learning to grasp from 50k tries and
  700 robot hours.
\newblock In \emph{IEEE Conference on Robotics and Automation (ICRA)}, 2016.

\bibitem[Pirsiavash \& Ramanan(2014)Pirsiavash and Ramanan]{Pirsiavash2014}
Hamed Pirsiavash and Deva Ramanan.
\newblock Parsing videos of actions with segmental grammars.
\newblock In \emph{IEEE Conference on Computer Vision and Pattern Recognition
  (CVPR)}, 2014.

\bibitem[Rusu et~al.(2016)Rusu, Colmenarejo, Gulcehre, Desjardins, Kirkpatrick,
  Pascanu, Mnih, Kavukcuoglu, and Hadsell]{Rusu2016}
Andrei~A Rusu, Sergio~Gomez Colmenarejo, Caglar Gulcehre, Guillaume Desjardins,
  James Kirkpatrick, Razvan Pascanu, Volodymyr Mnih, Koray Kavukcuoglu, and
  Raia Hadsell.
\newblock Policy distillation.
\newblock In \emph{International Conference on Learning Representations
  (ICLR)}, 2016.

\bibitem[Si et~al.(2011)Si, Pei, Yao, and Zhu]{Si2011}
Zhangzhang Si, Mingtao Pei, Benjamin Yao, and Song-Chun Zhu.
\newblock Unsupervised learning of event and-or grammar and semantics from
  video.
\newblock In \emph{IEEE International Conference on Computer Vision (ICCV)},
  2011.

\bibitem[Silver et~al.(2016)Silver, Huang, Maddison, Guez, Sifre, Driessche,
  and et~al.]{Silver2016}
David Silver, Aja Huang, Chris~J. Maddison, Arthur Guez, Laurent Sifre, George
  Van~Den Driessche, and Julian~Schrittwieser et~al.
\newblock Mastering the game of go with deep neural networks and tree search.
\newblock \emph{Nature}, 529\penalty0 (7587):\penalty0 484--489, 2016.

\bibitem[Silver et~al.(2017)Silver, Schrittwieser, Simonyan, Antonoglou, Huang,
  Guez, Hubert, Baker, Lai, Bolton, Chen, Lillicrap, Hui, Sifre, Driessche,
  Graepel, and Hassabis]{Silver2017}
David Silver, Julian Schrittwieser, Karen Simonyan, Ioannis Antonoglou, Aja
  Huang, Arthur Guez, Thomas Hubert, Lucas Baker, Matthew Lai, Adrian Bolton,
  Yutian Chen, Timothy Lillicrap, Fan Hui, Laurent Sifre, George Van~Den
  Driessche, Thore Graepel, and Demis Hassabis.
\newblock Mastering the game of go without human knowledge.
\newblock \emph{Nature}, 550\penalty0 (7676):\penalty0 354--359, 2017.

\bibitem[Su et~al.(2017)Su, Budzianowski, Ultes, Gasic, and Young]{Su2017}
Pei-Hao Su, Pawel Budzianowski, Stefan Ultes, Milica Gasic, and Steve Young.
\newblock Sample-efficient actor-critic reinforcement learning with supervised
  data for dialogue management.
\newblock In \emph{The 18th Annual SIGdial Meeting on Discourse and Dialogue
  (SIGDIAL)}, 2017.

\bibitem[Sutton et~al.(1999)Sutton, Precup, and Singh]{Sutton1999}
Richard~S. Sutton, Doina Precup, and Satinder Singh.
\newblock Between mdps anad semi-mdps: A framework fro temporal abstraction in
  reinforcement learning.
\newblock \emph{Artificial Intelligence}, 112:\penalty0 181--211, 1999.

\bibitem[Teh et~al.(2017)Teh, Bapst, Czarnecki, Quan, Kirkpatrick, Hadsell,
  Heess, and Pascanu]{Teh2017}
Yee~Whye Teh, Victor Bapst, Wojciech~Marian Czarnecki, John Quan, James
  Kirkpatrick, Raia Hadsell, Nicolas Heess, and Razvan Pascanu.
\newblock Distral: Robust multitask reinforcement learning.
\newblock In \emph{International Conference on Learning Representations
  (ICLR)}, 2017.

\bibitem[Tessler et~al.(2017)Tessler, Givony, Zahavy, Mankowitz, and
  Mannor]{Tessler2017}
Chen Tessler, Shahar Givony, Tom Zahavy, Daniel~J. Mankowitz, and Shie Mannor.
\newblock A deep hierarchical approach to lifelong learning in minecraft.
\newblock In \emph{AAAI Conference on Artificial Intelligence (AAAI)}, 2017.

\bibitem[Tieleman \& Hinto(2012)Tieleman and Hinto]{Tieleman2012}
Tijmen Tieleman and Geoffrey Hinto.
\newblock Lecture 6.5---rmsprop: Divide the gradient by a running average of
  its recent magnitude.
\newblock \emph{COURSERA: Neural Networks for Machine Learning}, 2012.

\bibitem[Vezhnevets et~al.(2016)Vezhnevets, Mnih, Agapiou, Osindero, Graves,
  Vinyals, and Kavukcuoglu]{Vezhnevets2016}
Alexander~(Sasha) Vezhnevets, Volodymyr Mnih, John Agapiou, Simon Osindero,
  Alex Graves, Orial Vinyals, and Koray Kavukcuoglu.
\newblock Strategic attentive writer for learning macro-actions.
\newblock In \emph{Advances in Neural Information Processing Systems (NIPS)},
  2016.

\bibitem[Wang et~al.(2017)Wang, Bapst, Heess, Mnih, Munos, Kavukcuoglu, and
  de~Freitas]{Wang2017}
Ziyu Wang, Victor Bapst, Nicolas Heess, Volodymyr Mnih, Remi Munos, Koray
  Kavukcuoglu, and Nando de~Freitas.
\newblock Sample efficient actor-critic with experience replay.
\newblock In \emph{International Conference on Learning Representations
  (ICLR)}, 2017.

\end{thebibliography}
\bibliographystyle{iclr2018_conference}

\newpage
\appendix
\section{Pseudo Code of Our Algorithms}\label{sec:app_algorithms}

\begin{algorithm}[h!] \scriptsize
\caption{RUN($k$, $g$)}
\label{alg:rollout}
\begin{algorithmic}[1]
\small

\INPUT Policy level $k$, task $g \in \mathcal{G}_k$
\OUTPUT Episode trajectory $\Gamma$ at the top level policy

\State $t \leftarrow 0$
\State $\Gamma = \emptyset$
\State Get initial state $s_0$

\Repeat
    \If {k == 1}
    	\State Sample $a_t \sim \pi_k(\cdot | s_t, g)$ and execute $a_t$
	\State Get current state $s_{t+1}$
	\State $r_t \leftarrow R(s_{t+1}, g)$
	\State Add $\langle s_t, a_t, r_t, \pi_k(\cdot | s_t, g), g \rangle$ to $\Gamma$
    \Else
    	\State Sample $e_t$ and $g^\prime_t$ as in Section~\ref{sec:grammar} for using STG as guidance
        \State Sample $a_t \sim \pi_{k}^\text{aug}(\cdot | s_t, g)$
	\If{$e_t = 0$}
		\State // \textit{Execute base policy $\pi_{k-1}$ by giving instruction $g^\prime_t$}
        \State RUN($k-1$, $g^\prime_t$)
	\Else
		\State Execute $a_t$
	\EndIf
	\State Get current state $s_{t+1}$
	\State $r_t \leftarrow R(s_{t+1}, g)$
	\State Add $\langle s_t, e_t, g^\prime_t, a_t, r_t, \pi^\text{sw}_k(\cdot | s_t), \pi_k^\text{inst}(\cdot | s_t, g), \pi_{k}^\text{aug}(\cdot | s_t, g), g \rangle$ to $\Gamma$ 
    \EndIf

    \State $t \leftarrow t + 1$
    
\Until{$t > T$ or $r_t \neq 0$}
\end{algorithmic}
\end{algorithm}

\begin{algorithm}[h!] \scriptsize
\caption{Learning global policy and STG at stage $k > 0$}
\label{alg:learning}
\begin{algorithmic}[1]
\small

\State Specify $\lambda$, maximum training iterations $N$, alternating update rotation frequency $M$, and reward threshold $R_\text{min}$
\State Initialize total replay memory $D \leftarrow \emptyset$ and its subset for positive episodes $D_+ \leftarrow \emptyset$
\State Initialize current iteration id $i \leftarrow 0$ and set current updating term to be $\tau \leftarrow 1$
\State Initialize parameters of policies and value functions $\Theta = \langle \theta^\text{sw}, \theta^\text{inst}, \theta^\text{aug}, \theta_v, \theta_v^\text{sw} \rangle$
\State Initialize distributions of the STG as uniform distributions

\Repeat
	\State Determine current learning phase by comparing average rewards of tasks in $\mathcal{G}_{k-1}$ with $R_\text{min}$
    \If {in curriculum learning phase 1}
    	\State Sample a task $g$ from base task set $\mathcal{G}_{k-1}$
    \Else
    	\State Sample a task $g$ from global task set $\mathcal{G}_k$
   	\EndIf
    \State //Run an episode
    \State $\Gamma \leftarrow$ RUN($k$, $g$)
	\State $D \leftarrow D \cup \Gamma$
    \If {the maximum reward in $\Gamma$ is +1}
    	\State $D_+ \leftarrow D_+ \cup \Gamma$
        \State Re-estimate the distributions of the STG based on updated $D_+$ by MLE
    \EndIf
    \State Sample $n \sim \text{Possion}(\lambda)$
    \For {$j \in \{1, \cdots, n\}$}
    	\State Sample a mini-batch $S$ from $D$
        \State Update $\Theta$ based on (\ref{eq:value_gradient}) and the $\tau$-th term in (\ref{eq:policy_gradient})
   		\State $i \leftarrow i + 1$
        \If {$i \% M = 0$}
        	\State $\tau \leftarrow \tau \% 3 + 1$
        \EndIf
    \EndFor
\Until{$i \geq N$}
\end{algorithmic}
\end{algorithm}

\section{Architectures of Modules}\label{sec:app_details}
The architecture designs of all modules in our model shown in Figure~\ref{fig:archs} are as follows:

\textbf{Visual Encoder} extracts feature maps from an input RGB frame with the size of $84\times 84$ through three convolutional layers: i) the first layer has 32 filters with kernel size of $8 \times 8$ and stride of $4$; ii) the second layer has 64 filters with kernel size of $4 \times 4$ and stride of 2; iii) the last layer includes 64 filters with kernel size of $3 \times 3$ and stride of 1. The feature maps are flatten into a 3136-dim vector. We reduce the dimension of this vector to 256 by a fully connected (FC) layer resulting a 256-dim visual feature as the final output of this module.

\textbf{Instruction Encoder} first embeds each word into a 128-dim vector and combines them into a single vector by bag-of-words (BOW). Thus the output of this module is a 128-dim vector.

\textbf{Fusion} layer simply concatenates the encoded visual and language representations together and outputs 384-dim fused representation. We then feed this 384-dim vector into an \textbf{LSTM} with 256 hidden units. The hidden layer output of the LSTM is served as the input of all policy modules and value function modules.

\textbf{Switch Policy} module has a FC layer with output dimension of 2 and a softmax activation to get $\pi_k^\text{sw}(e | s, g)$. \textbf{Instruction Policy} module has two separate FC layers, both of which are activated by softmax to output the distribution of \textit{skill}, $p_k^\text{skill}(u_\text{skill} | s, g)$, and the distribution of \textit{item}, $p_k^\text{item}(u_\text{item} | s, g)$, respectively. \textbf{Augmented Policy} module outputs $\pi_\text{aug}(a | s, g)$ also through a FC layer and softmax activation. The two \textbf{Value Function} modules, $V(s, g)$ and $V^\text{sw}(s, e, g)$, all have a scalar output through a FC layer.

Finally, the \textbf{Selector} module selects the action sampled from \textbf{Augmented Policy} module or \textbf{Base Policy} module based on the switching decision sampled from the \text{Switch Policy} module.

\section{Composed Hierarchical Plans}\label{sec:app_viz}
\begin{figure}[t!]
\centering
\includegraphics[trim={0 0 0 0},clip,width=0.9\linewidth]{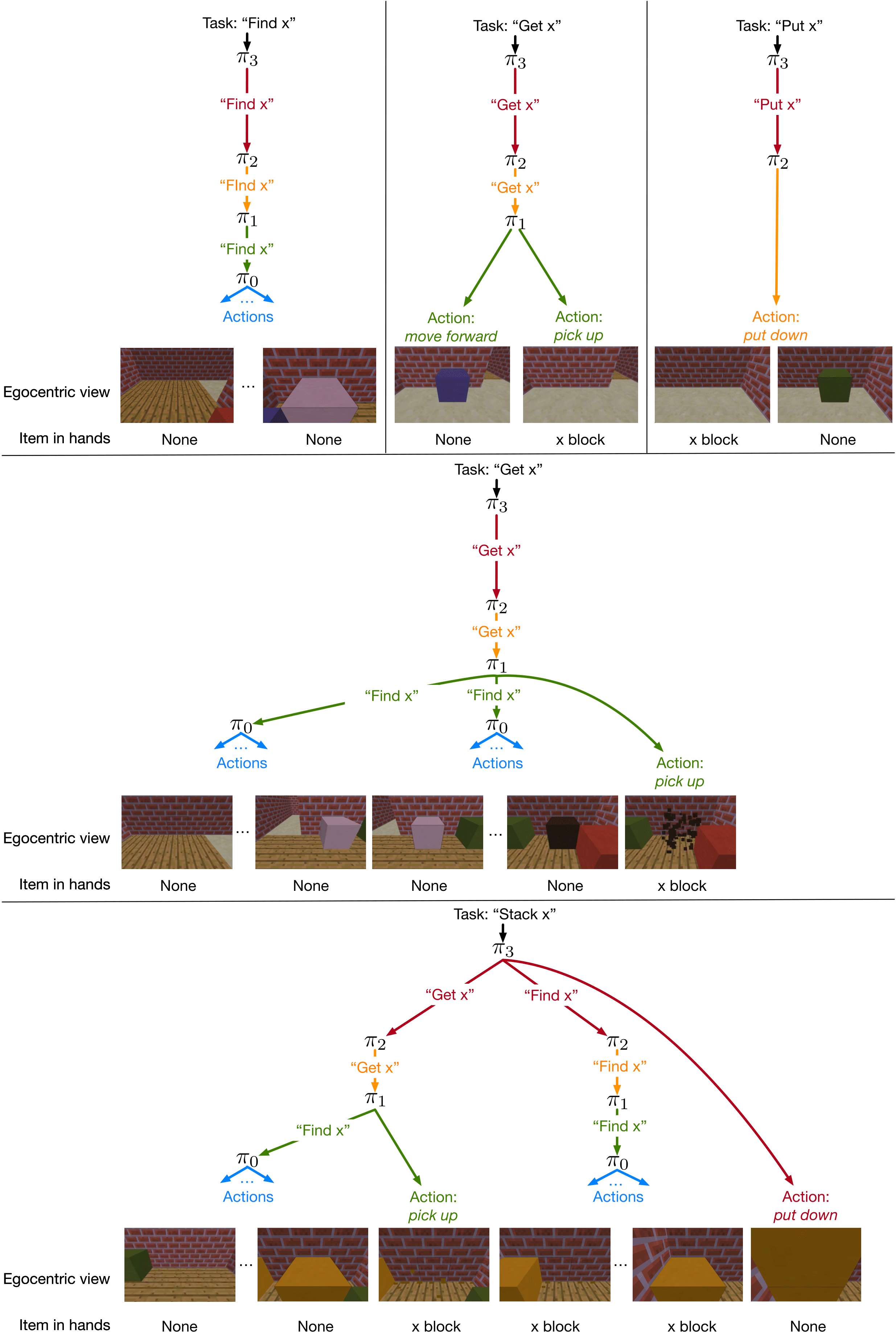}
\caption{Samples of typical hierarchical plans for different tasks composed by our global policies. Note that all tasks must start from the top-level policy. The branches are ordered from left to right in time indicating consecutive steps carried out by a policy. We also show the egocentric view and the item in hands at critical moments for a real episode example.}
\label{fig:viz_policies}
\end{figure}

Figure~\ref{fig:viz_policies} shows several plans for different tasks composed by executing our hierarchical policies.

\end{document}